\documentclass[sigconf]{acmart}

\usepackage{amsmath}
\usepackage{booktabs}
\usepackage{multirow}
\usepackage{pifont}
\newcommand{\cmark}{\ding{51}}
\newcommand{\xmark}{\ding{55}}
\graphicspath{{figures/}}
\setcopyright{none}
\renewcommand\footnotetextcopyrightpermission[1]{}
\acmConference[BIOKDD '26]{The 25th International Workshop on Data Mining in Bioinformatics}{August 2026}{Jeju, Republic of Korea}
\acmYear{2026}
\copyrightyear{2026}

\begin{document}

\title{Predicting Therapeutic Outcome via Aligning Patient-Specific Knowledge Graph and Gene-Level Perturbation Representations}

\author{Dongmin Bang}
\authornote{These authors contributed equally to this work.}
\affiliation{\institution{Interdisciplinary Program in Bioinformatics, Seoul National University}\city{Seoul}\postcode{08826}\country{Republic of Korea}}
\affiliation{\institution{AIGENDRUG Co., Ltd.}\city{Seoul}\postcode{08758}\country{Republic of Korea}}

\author{Sugyun An}
\authornotemark[1]
\affiliation{\institution{AIGENDRUG Co., Ltd.}\city{Seoul}\postcode{08758}\country{Republic of Korea}}

\author{Inyoung Sung}
\affiliation{\institution{BK21 FOUR Intelligence Computing, Seoul National University}\city{Seoul}\postcode{08826}\country{Republic of Korea}}

\author{Ilho Yun}
\affiliation{\institution{Interdisciplinary Program in Artificial Intelligence, Seoul National University}\city{Seoul}\postcode{08826}\country{Republic of Korea}}

\author{Sun Kim}
\affiliation{\institution{Interdisciplinary Program in Bioinformatics, Seoul National University}\city{Seoul}\postcode{08826}\country{Republic of Korea}}
\affiliation{\institution{AIGENDRUG Co., Ltd.}\city{Seoul}\postcode{08758}\country{Republic of Korea}}
\affiliation{\institution{Interdisciplinary Program in Artificial Intelligence, Seoul National University}\city{Seoul}\postcode{08826}\country{Republic of Korea}}

\author{Sangseon Lee}
\authornote{Corresponding author.}
\affiliation{\institution{Department of Artificial Intelligence, Inha University}\city{Incheon}\postcode{22212}\country{Republic of Korea}}
\email{ss.lee@inha.ac.kr}

\begin{abstract}

Accurate prediction of patient-specific therapeutic response from pre-treatment transcriptomes is hindered by the scarcity of matched clinical response labels and post-treatment molecular profiles. Preclinical transfer-learning models can simulate drug-induced expression changes but are often hard to interpret and unstable, whereas knowledge-graph methods provide mechanistic context yet remain static and fail to capture drug-induced transcriptomic perturbation dynamics. We propose PREDIKTOR, a patient-centered multi-view framework that aligns a personalized network view with a transferable transcriptomic perturbation view to predict clinical drug response. For each patient, we construct an individualized gene regulatory network from tumor expression using DysRegNet and augment it with drug–target links from DrugBank; a graph neural encoder yields a drug-centric, mechanistically grounded embedding. In parallel, a frozen condition-specific gene–gene attention model pretrained on LINCS L1000 generates a simulated post-perturbation transcriptomic profile for the same patient–drug pair. We align the two views in a shared latent space via a CLIP-style contrastive objective with drug-context hard negatives, then concatenate the representations for end-to-end response classification. On TCGA, PREDIKTOR consistently outperforms state-of-the-art baselines under patient-, drug-, and tissue-split evaluations, and transfers zero-shot to the I-SPY2 trial, improving AUROC by 5.6\% over competing methods. The aligned embeddings yield stable gene and pathway attributions that recover known mechanisms, supporting actionable and interpretable precision oncology.
\end{abstract}



\maketitle
\section{1. Introduction}
Personalized drug response prediction, which aims to forecast how an individual patient will respond to a given therapy from their molecular profile, remains a central challenge in precision oncology \cite{gameiro2018precision, evans1999pharmacogenomics}. Clinically actionable prediction requires models that move beyond correlative associations to capture how drug induced perturbations reprogram molecular states within a heterogeneous and patient specific cellular context \cite{adam2020machine}. In practice, however, large patient cohorts rarely provide matched patient level drug response labels and post treatment molecular measurements at scale, even when baseline tumor transcriptomes are available, as exemplified by The Cancer Genome Atlas (TCGA) \cite{weinstein2013cancer}. As a result, most existing approaches have relied on large preclinical resources, including the Genomics of Drug Sensitivity in Cancer (GDSC) \cite{iorio2016landscape} and the Cancer Cell Line Encyclopedia (CCLE) \cite{barretina2012cancer}, to train drug response prediction models \cite{samal2022opportunities}.

\begin{figure}[t]
    \centering
    \includegraphics[width=\linewidth]{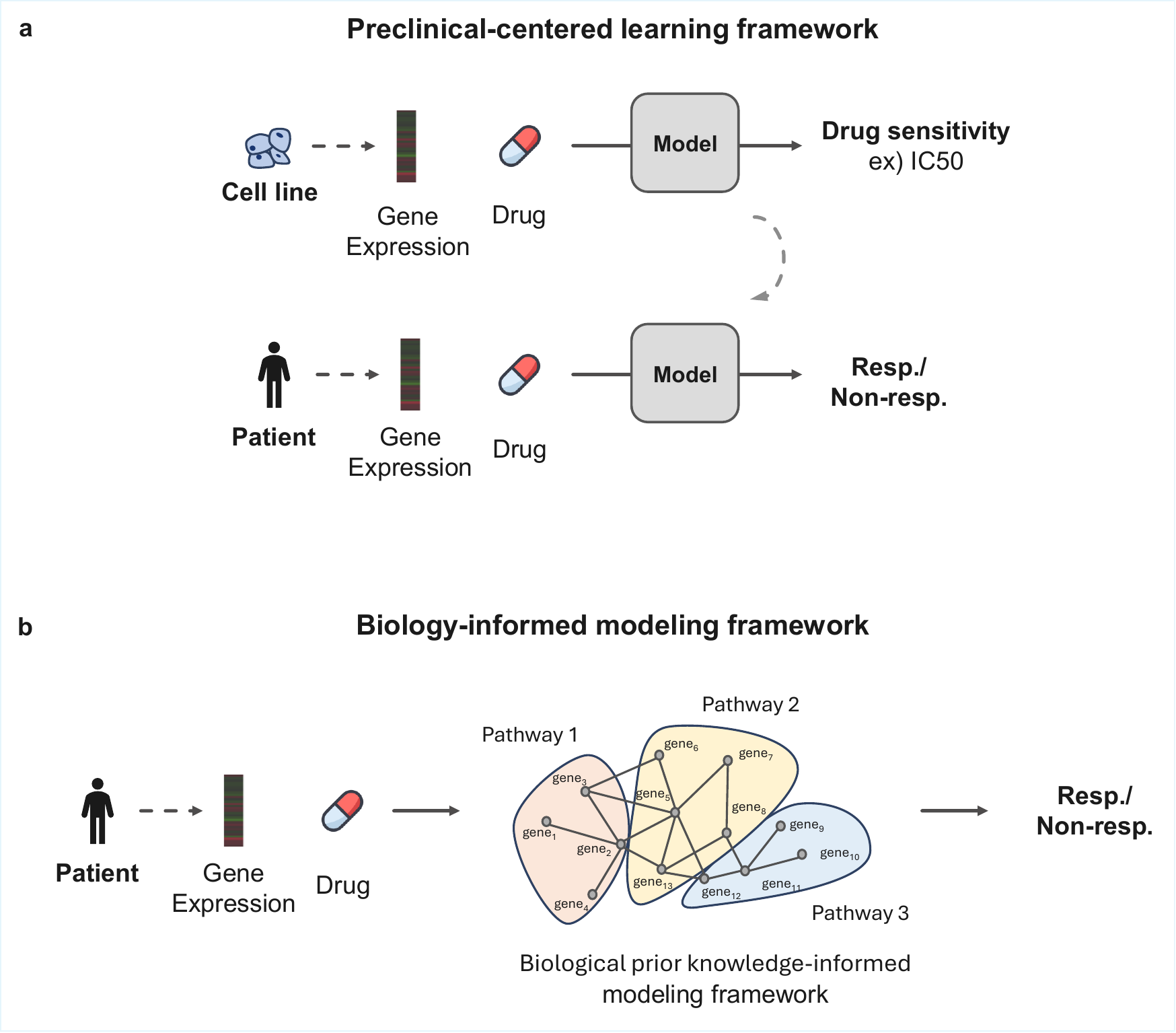}
    \caption{\normalfont\textbf{Abstract visualization of existing drug response prediction frameworks.} \textbf{a} Preclinical-centered learning paradigm, where models are trained on cancer cell line molecular profiles and drug sensitivity readouts (e.g., IC50) and directly transferred to patient-level drug response prediction. \textbf{b} Biology-informed modeling paradigm, which incorporates prior biological knowledge (e.g., pathways or gene regulatory networks) or perturbation-based data (e.g., LINCS L1000) to learn more transferable and biologically informed representations for patient-level prediction.}
    \label{fig:exising_works}
\end{figure}

Early approaches consequently adopted preclinical centered learning strategies, associating cell line molecular features with drug sensitivity readouts (e.g, IC50) (Figure~\ref{fig:exising_works}a). Representative examples include DeepCDR \cite{liu2020deepcdr}, a hybrid graph convolutional model that integrates cell line multi-omics data with drug structural information, and DeepTTA \cite{jiang2022deeptta}, which employs Transformer-based representations of drug SMILES sequences combined with transcriptomic features to estimate IC50 values in cancer cell lines. Although these approaches have demonstrated strong performance in preclinical benchmarks, their clinical translation to patient-level drug response prediction has remained limited \cite{sharifi2021drug}. A key reason is that such models are optimized to learn static associations from preclinical molecular profiles to drug response, rather than representing drug-induced dynamics and patient-specific regulatory context that determine clinical efficacy.

To improve robustness and biological plausibility, recent studies have incorporated biological pathway or perturbation-informed representations (Figure~\ref{fig:exising_works}b). For example, DRPreter introduces pathway-informed graph embeddings to enhance interpretability \cite{shin2022drpreter}, while Precily integrates pathway activity scores with drug descriptors and extends evaluation to patient cohorts from TCGA \cite{chawla2022precily}. More recently, CSG\textsuperscript{2}A leverages the library of integrated network-based cellular signatures (LINCS) L1000 perturbation profiles \cite{duan2016l1000cds2} to learn transferable gene–gene interaction patterns that can be applied to patient transcriptomes \cite{bang2024transfer}. While these approaches provide more biologically informed representations and enable limited translation to patient-level drug response prediction, they are still primarily driven by representations learned from preclinical data and do not fully leverage patient-level therapeutic data that reflect complex, tissue-level, and patient-specific biological heterogeneity.

\begin{figure*}[t]
    \centering
    \includegraphics[width=\textwidth]{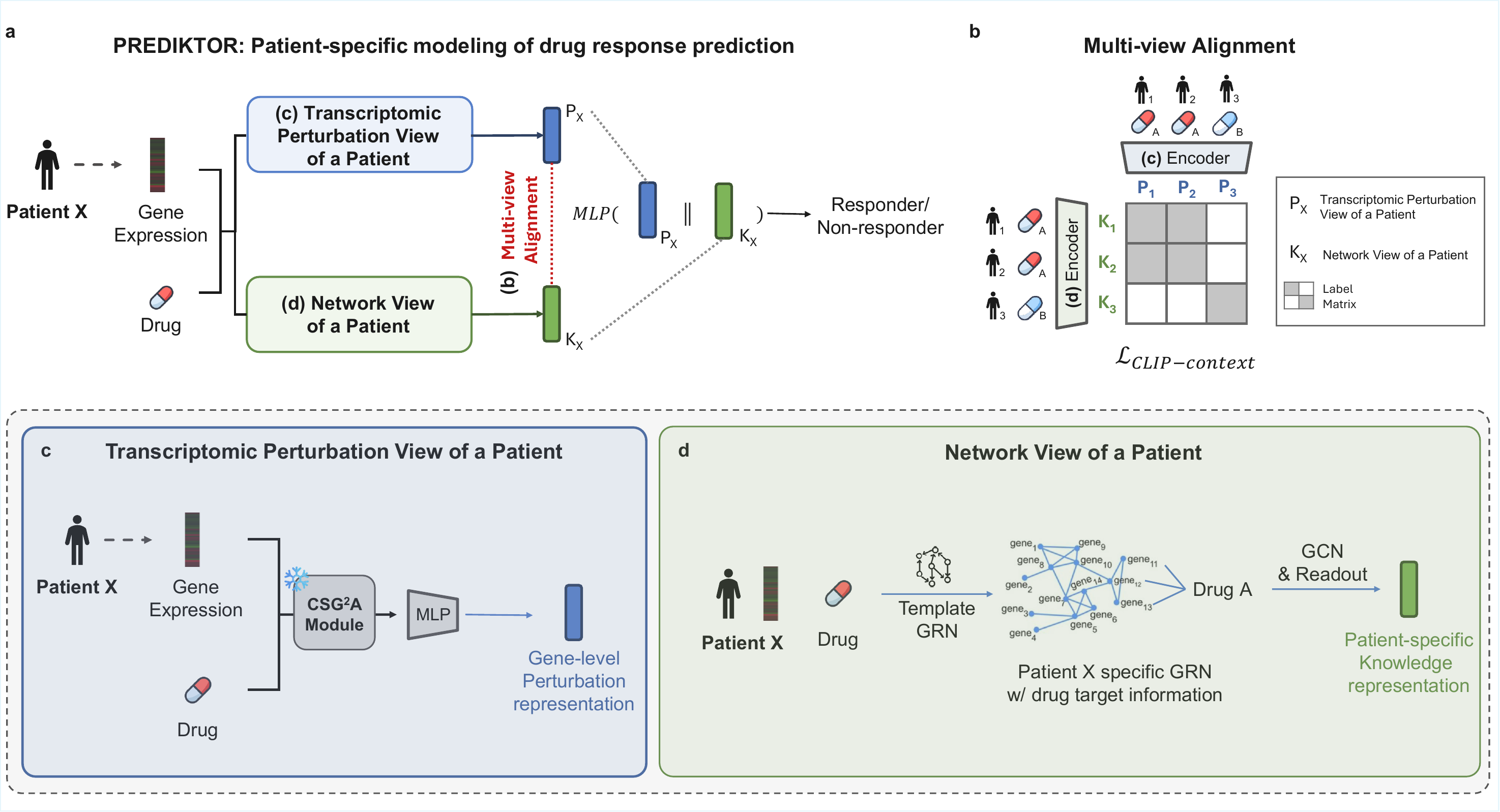}
    \caption{\normalfont Overview of \textsc{PREDIKTOR}: 
    \textbf{P}ersonalized drug
    \textbf{R}esponse
    \textbf{E}stimation via
    \textbf{D}ual
    \textbf{I}ntegration of
    \textbf{K}nowledge graph and
    \textbf{T}ransferable
    \textbf{O}mics-driven
    \textbf{R}epresentations. \textbf{a} Overall architecture of PREDIKTOR, a patient-centered framework that takes patient transcriptomes as the primary input and predicts patient-specific drug responses. \textbf{b} Contrastive multi-view alignment module, inspired by CLIP, which aligns complementary representations from transcriptomic perturbation and network views into a unified patient–drug embedding. \textbf{c} Transcriptomic perturbation view, where a transferable perturbation encoder is trained on large-scale drug perturbation experiments to model generalizable drug-induced transcriptional changes, providing an approximation of post-treatment molecular states in the absence of patient-level perturbation data. \textbf{d} Network view, where a patient-specific knowledge graph encoder contextualizes prior biological networks with each patient’s transcriptomic state to construct personalized gene regulatory representations. The unified multi-view embedding is subsequently used for end-to-end prediction of patient-specific drug responses.    
    }
    \label{fig:overview}
\end{figure*}

To address these limitations, we introduce PREDIKTOR, Personalized Drug Response Estimation via Dual Integration of Knowledge Graph and Transferable Omics-driven Representations, a novel patient-centered framework that uses patient transcriptomes as the core input for predicting patient specific drug responses (Figure~\ref{fig:overview}a). PREDIKTOR addresses the central bottleneck, namely the absence of patient specific post perturbation molecular measurements, by integrating multi-views of each patient-drug pair. In a first \textit{transcriptomic perturbation view} (Figure~\ref{fig:overview}c), a transferable perturbation encoder learns generalizable patterns of drug induced transcriptional change from large scale perturbation experiments, providing an approximation for post treatment transcriptional reorganization when patient level perturbation profiles are unavailable. In a second \textit{network view} (Figure~\ref{fig:overview}d), a patient specific knowledge graph encoder contextualizes prior biological networks with each patient’s transcriptomic state to represent a personalized gene regulatory network. We align these multi-view representations with a contrastive objective inspired by CLIP \cite{radford2021clip} (Figure~\ref{fig:overview}b), learning a unified patient drug embedding that couples dynamic perturbation signatures with a patient specific mechanistic context. The unified multi-view representation is subsequently used for the end-to-end prediction of the patient-specific drug response.

Together, these components yield a gene expression–based drug response predictor that combines strong generalization with human-interpretable, mechanism-level explanations. On TCGA, PREDIKTOR consistently outperforms state-of-the-art baselines across diverse and clinically relevant evaluation protocols, including patient-split, drug-split, and tissue-split settings, and shows robust gains across most metrics. We further demonstrate generalizability on an external cohort, I-SPY2 \cite{barker2009ispy}, where PREDIKTOR improves AUROC by 5.6\% relative to competing approaches. Collectively, these results establish PREDIKTOR as a principled patient-centric framework that bridges transferable perturbation knowledge with individualized tumor biology, advancing the clinical applicability of transcriptome-driven precision oncology models.

\section{2. Materials and methods}
\subsection{2.1. Problem formulation}
We define the problem of patient drug response prediction as following:
Let \( x_p \in \mathbb{R}^G \) denote the gene expression profile of a patient \( p \), where \( G \) is the number of profiled genes, and let \( d \in \mathcal{D} \) be a candidate drug with molecular structure represented by its SMILES encoding. 
The goal is to learn a function $f: (x_p, d) \mapsto \hat{y}_{p,d} \in [0, 1] $ where \( \hat{y}_{p,d} \) is the predicted probability of a positive therapeutic response when drug \( d \) is administered to patient \( p \). The true drug response label \( y_{p,d} \in \{0,1\} \) is binary, based on clinical outcomes.
\subsection{2.2. Model overview}
To predict patient-specific therapeutic response with biological interpretability, we propose PREDIKTOR, a multi-view framework that fuses two complementary representations: (i) a knowledge graph–based \textit{Network View} capturing patient-specific biological context, and (ii) a transfer-learned \textit{Transcriptomic Perturbation View} capturing perturbation signatures under drug treatment.  The two view encoders produce embeddings that are concatenated and fed to an MLP for the patient-specific drug response prediction. Since both views originate from the same patient, we align their embeddings as closely as possible before feeding them into the MLP. 
\begin{equation}
\begin{aligned}
f(x_p, d) = \text{MLP}(&
\underbrace{f_\text{GCN}(\mathcal{G}_p, d_k, x_p)}_{\text{Knowledge Graph-based Network View}} \\
&\| \;
\underbrace{f_\text{CSG2A}(x_p, d_s)}_{\text{Transcriptomic Perturbation View}}
)
\end{aligned}
\end{equation}

where \( \mathcal{G}_p \) is a patient-specific graph, $d_k$ represents drugs via their therapeutic targets, $d_s$ represents a drug by its chemical structure, \( \| \) denotes concatenation, and each encoder is described below. 
\subsection{2.3. Network view: patient-specific knowledge graph encoding}
\subsubsection{2.3.1. Gene Regulatory Network (GRN) construction}
Patient-specific GRNs were constructed using DysRegNet\cite{kersting2024dysregnet}, a gene expression-based framework that detects dysregulated transcription factor (TF)-target gene (TG) interactions at individual level. Briefly, DysRegNet fits a baseline linear model using control samples from matched healthy tissues in TCGA, with TF expression as predictors and TG expression as responses. For each patient sample, it then quantifies deviations from this baseline (via residual-based tests) to identify significantly dysregulated TF–TG interactions. We included edges passing Benjamini–Hochberg FDR $< 0.05$. Candidate TF–TG pairs were restricted to experimentally supported interactions from HTRIdb \cite{bovolenta2012htridb} to ensure biological plausibility.

\subsubsection{2.3.2. Integration of drug-gene interactions}
Drug-specific gene interactions were integrated from DrugBank\cite{knox2024drugbank}, a curated database of approved drug information. Drugs were mapped to target proteins, constrained to the preselected 7,800-gene universe. These were added as external nodes to the previously homogeneous gene-only GRN, forming a heterogeneous graph where drug nodes connect to target genes via directed edges. Consequently, for each patient \( p \), we construct a patient-specific biomedical knowledge graph \( \mathcal{G}_p = (V_p, E_p) \), where \( V_p \) includes gene nodes and the drug node, and \( E_p \) consists of (i) dysregulated TF-target gene interactions inferred from patient transcriptomics via DysRegNet, and (ii) drug-target links from DrugBank.

\subsubsection{2.3.3. Graph neural network for learning patient-specific drug representation}
The GCN\cite{kipf2016semi} stream processes each patient’s heterogeneous knowledge graph. Initially, the graph consists of gene nodes only, forming a directed homogeneous GRN. Upon integrating the drug node via known target interactions, the graph becomes heterogeneous. Node embeddings, \( h_v^{(0)} \in \mathbb{R}^{Dw} \), for node \( v \in V_p \) are initialized using DREAMwalk \cite{bang2023biomedical}, a semantic random-walk-based embedding technique accounting for biomedical graph heterogeneity.
We employ a two-layer Graph Convolutional Network (GCN) architecture with ReLU activation and hidden dimensionality of 128. Embeddings are updated as
\[ H^{(l+1)} = \sigma(\tilde{D}^{-1/2}\tilde{A}\tilde{D}^{-1/2}H^{(l)}W^{(l)}) \]
where \(\tilde{A}\) is the adjacency matrix with self-loops, \(\tilde{D}\) the corresponding degree matrix, and \(W^{(l)}\) learnable parameters. Node embeddings had dimensions of 128, with ReLU activation functions applied after each convolutional layer.
The final representation for the drug node \( h_d^{(L)} \) is sum-pooled over its local neighborhood. The resulting drug-centric graph embedding is then concatenated with the patient transcriptomic representation  \( e_p \) to form the knowledge-graph-based network view,

$f_{GCN}(\mathcal{G}_p, d_k, x_p) = z^{\text{graph}}_{p,d_k} = \left[ h_d^{(L)} || e_p \right]
\in \mathbb R^{n_\text{graph}+ n_\text{tx}}$, where  \( e_p \in \mathbb{R}^{n_{\text{tx}}} \) denotes the layer-normalized \( x_p \), and $n_\text{grpah} = n_\text{tx} = 128$.

\subsection{2.4. Transcriptomic perturbation view: transfer learning from large-scale perturbation data}
Before training PREDIKTOR, the condition-specific Gene-Gene Attention (CSG$^2$A) module is pretrained on the LINCS L1000 dataset, which includes transcriptional responses of cell lines to a wide range of chemical perturbagens \cite{bang2024transfer}. Preprocessing included normalization, filtering for consistent treatment times, and use of the core 978 landmark genes.
CSG$^2$A uses a self-attention mechanism over gene-gene pairs constrained by a reference PPI network from STRING database \cite{szklarczyk2023string}. Both the Query and Key inputs to the self-attention layers are the sum of MAT-encoded drug embeddings \cite{maziarka2020molecule} and one-hot gene identifiers. The attention matrix $A=[\alpha_{ij}]\in \mathbb R^{n_\text{gene}\times n_\text{gene}}$, where $n_\text{gene} = 978$, captures the conditional gene-gene interactions, with $\alpha_{ij}$ representing the pretrained influence of gene \( j \) on gene \( i\) under drug  $d_s$ perturbation.

During the end-to-end learning of PREDIKTOR, the pretrained weights are frozen to preserve the learned attention scores. The perturbed gene expression for gene $i$ of patient $p$ with drug $d_s$ is predicted by aggregating attention-weighted baseline expressions and passing them through an MLP: $\hat{x}_p^{\text{perturbed}}[i] = \text{MLP}\left(\sum_j \alpha_{ij}\cdot x_p[j]\right)$. The predicted perturbation values for all genes are then stacked to form a transcriptomic perturbation view, $f_{CSG2A}(x_p, d_s) = z^{\text{perturb}}_{p,d_s}\in \mathbb{R}^{n_\text{gene}} $\ . This vector is then used without additional transformation as the output of the perturbation attention stream.

\subsection{2.5. Training strategy and objective functions}
\subsubsection{2.5.1. Contrastive alignment of patient representations}
We encourage the Network View and Transcriptomic Perturbation View to align by adding a symmetric CLIP-style contrastive loss \cite{radford2021clip}. For each patient–drug pair, the network view representation $z^{\text{graph}}_{p,d_k}\in \mathbb{R}^{n_\text{graph}+n_\text{tx}}$, and the perturbation view representation $z^{\text{perturb}}_{p,d_s}\in \mathbb{R}^{n_\text{gene}}$  are linearly projected into a shared space, $z^G_{p,d} = W_G\, z^{\text{graph}}_{p,d_k}$, $z^P_{p,d} = W_P\, z^{\text{perturb}}_{p,d_s}$, where $W_G \in \mathbb{R}^{n_d \times( n_{\text{graph}}+n_{\text{tx}})}$ and $W_P \in \mathbb{R}^{n_d \times n_{\text{gene}}} $ are projection matrices.
Then, these multi-view representations are optimized to maximize cosine similarity for matched pairs while separating mismatched pairs within a mini-batch:
\[ \begin{aligned} \mathcal{L}_{\text{CLIP}} &= \frac{1}{2N} \sum_{i=1}^N \Bigg[ -\log \frac{\exp(\mathrm{sim}(z^G_i, z^P_i)/\tau)} {\sum_{j=1}^N \exp(\mathrm{sim}(z^G_i, z^P_j)/\tau)} \\ &\qquad\qquad\qquad -\log \frac{\exp(\mathrm{sim}(z^P_i, z^G_i)/\tau)} {\sum_{j=1}^N \exp(\mathrm{sim}(z^P_i, z^G_j)/\tau)} \Bigg]. \end{aligned} \]
where \( \text{sim}(a, b) = \frac{a^\top b}{\|a\|\|b\|} \) denotes cosine similarity, N is the number of patient-drug pairs in a given batch, and \( \tau \) is a learnable temperature parameter.

\paragraph{Hard negative selection based on common drugs:}
Effective contrastive learning requires informative hard negatives to establish precise decision boundaries in shared embedding spaces. Our experiments demonstrate that non-drug-sharing pairs, where patient representations are coupled with perturbations from different drugs, provide the most effective negatives. We therefore adopt this strategy, denoted as $\mathcal{L}_{\text{CLIP-context}}$, which yields superior performance by using pharmacologically plausible yet mechanistically distinct negatives, thereby forcing the model to capture drug-specific context. In contrast, selecting negatives based on tissue types or supervised labels caused performance to drop. These approaches introduce semantic noise by conflating biological variability with treatment effects, which ultimately prevents the model from forming accurate intermodal alignments.

\subsubsection{2.5.2. Prediction head for drug response prediction}

The final input to the prediction head is the concatenation of two feature vectors: (1) the 256-dimensional knowledge graph-based network view $z^{\text{graph}}_{p,d_k}$ from the GCN and (2) the 978-dimensional simulated transcriptomic perturbation veiw $z^{\text{perturb}}_{p,d_s}$ from CSG$^2$A. No additional attention or gating is used between the two representations.
These concatenated vectors are fed into a two-layer Multilayer Perceptron (MLP) consisting of a 128-unit hidden layer with ReLU activation, followed by dropout (rate = 0.1), Layer Normalization, and a sigmoid output layer predicting drug response, which is then evaluated using the binary cross-entropy (BCE) loss function.

\subsubsection{2.5.3. Objective loss function}
The final loss used during training is a weighted sum of the binary cross-entropy loss \( \mathcal{L}_{\text{BCE}} \) and the CLIP loss:
\[ \mathcal{L}_{\text{total}} = \mathcal{L}_{\text{BCE}} + \lambda_{\text{CLIP}} \mathcal{L}_{\text{CLIP-context}}. \]
Here, \( \lambda_{\text{CLIP}} \) is a trainable hyperparameter that controls the CLIP loss's contribution, initially set to 5. 
The model is trained end-to-end using the Adam optimizer (learning rate = 1e-4, batch size = 32). The CSG$^2$A and MAT modules are frozen throughout training. Training is performed using PyTorch on NVIDIA RTX 3090 GPUs, with early stopping based on validation loss value. Representative training and validation loss curves for the random-, drug-, and tissue-split settings are provided in Supplementary Figure~\ref{fig:supp_lossplot}.

\subsection{2.6. Dataset and evaluation setting}
\subsubsection{2.6.1. Datasets}
\paragraph{TCGA dataset:}
The primary dataset used in this study is sourced from The Cancer Genome Atlas (TCGA), accessed through the UCSC Xena Cancer Genome Browser (https://xena.ucsc.edu/). A total of 358 patients with clearly annotated clinical drug response data across 21 unique drugs were selected, resulting in 383 patient-drug response pairs. Among the 21 TCGA drugs, 18 overlap with compounds represented in the LINCS L1000 data used to pretrain CSG$^2$A. While this overlap may provide prior information about drug-induced transcriptional effects, LINCS L1000 does not contain TCGA clinical drug response labels.
Clinical drug responses were binarized (responders vs. non-responders) according to standardized response annotations provided by TCGA clinical metadata. Gene expression data were processed using RNA-seq expression values normalized as log2(TPM + 1). We filtered for gene overlap with the reference gene regulatory network (GRN) and biomedical knowledge graph (KG), resulting in a consistent gene set of 7,800 genes. To comprehensively evaluate generalization capabilities, we used three mutually exclusive data-splitting settings: (1) patient-split, (2) drug-split, and (3) tissue-split, ensuring that the corresponding patients/drugs/tissues are strictly non-overlapping between training and testing sets. Each split was evaluated using stratified 5-fold cross-validation with the train, validation, test ratio of 5:1:1 to minimize data leakage and maintain balanced class distributions for fair performance comparison.

\paragraph{External dataset:}
For external validation, we utilized the I-SPY2 dataset \cite{barker2009ispy}, comprising microarray expression profiles from pre-treatment breast cancer biopsies of 988 patients. A subset of 178 patients treated specifically with paclitaxel was used, with responses defined by pathological complete response (pCR): binary labeled as 1 (complete response) or 0 (no complete response). This dataset served exclusively as a zero-shot inference set without any retraining, testing the model's direct transferability to unseen clinical scenarios.

\subsubsection{2.6.2. Evaluation setting}
\paragraph{Comparison models:}
We benchmarked PREDIKTOR against a representative suite of state-of-the-art drug response prediction methods spanning the current biomedical AI landscape. The comparison models were grouped into three categories: (1) traditional and basic baselines, including established machine learning methods (e.g., XGBoost, Random Forest, and SVM) and fundamental deep learning architectures (e.g., DeepCDR \cite{liu2020deepcdr}, DeepTTA \cite{jiang2022deeptta}); (2) knowledge-guided models that explicitly incorporate biological priors, such as protein–protein interaction networks or knowledge graphs (e.g., DRPreter \cite{shin2022drpreter}, Precily \cite{chawla2022precily}); and (3) transfer learning–based attention models that leverage large-scale omics data to capture complex drug-induced effects (e.g., CSG$^2$A \cite{bang2024transfer}, GeneFormer \cite{theodoris2023geneformer}).

\paragraph{Evaluation metrics:}
Model performance was evaluated comprehensively using Area Under the Receiver Operating Characteristic Curve (AUROC), Area Under the Precision-Recall Curve (AUPRC), Accuracy, and F1 Score. Metrics were computed across folds and averaged, with variability assessed via repeated experiments through 5-fold cross-validation.

\section{3 Results}

\subsection{3.1 Performance of patient-specific knowledge-guided framework on TCGA dataset}

\subsubsection{3.1.1 Predictive performance across evaluation splits}

\begin{table*}[t]
\caption{\normalfont \textbf{Evaluation of predictive performance across diverse settings.} Patient drug response prediction performances of comparison models on three different split settings of the TCGA dataset. Mean and standard deviation of 5-fold cross validation are provided. Best performances are marked in bold and second-best are underlined.}
\resizebox{\textwidth}{!}{%
\begin{tabular}{l|rrr|rrr|rrr}
\toprule
 & \multicolumn{3}{c|}{\textbf{TCGA--Patient split}} & \multicolumn{3}{c|}{\textbf{TCGA--Drug split}} & \multicolumn{3}{c}{\textbf{TCGA--Tissue split}} \\
 & \multicolumn{1}{c}{AUROC} & \multicolumn{1}{c}{AUPRC} & \multicolumn{1}{c|}{ACC} & \multicolumn{1}{c}{AUROC} & \multicolumn{1}{c}{AUPRC} & \multicolumn{1}{c|}{ACC} & \multicolumn{1}{c}{AUROC} & \multicolumn{1}{c}{AUPRC} & \multicolumn{1}{c}{ACC} \\
 \midrule
SVM & 0.803 (0.0371) & 0.775 (0.0473) & 0.733 (0.0235) & 0.576 (0.0682) & \underline{0.558 (0.2350)} & 0.490 (0.2398) & 0.548 (0.0477) & 0.540 (0.2362) & 0.487 (0.1242) \\
Random Forest & 0.832 (0.0422) & 0.792 (0.0671) & \textbf{0.778 (0.0390)} & \underline{0.582 (0.0831)} & 0.551 (0.2528) & \underline{0.567 (0.1690)} & 0.534 (0.0362) & 0.531 (0.2658) & 0.526 (0.1658) \\
XGBoost & 0.818 (0.0567) & 0.776 (0.0717) & 0.770 (0.0493) & 0.570 (0.0877) & 0.541 (0.2726) & 0.553 (0.1888) & 0.527 (0.0622) & 0.535 (0.2556) & 0.554 (0.1260) \\
DeepCDR & 0.824 (0.0394) & 0.784 (0.0644) & 0.759 (0.0445) & 0.533 (0.0600) & 0.507 (0.2431) & 0.353 (0.2101) & 0.495 (0.1000) & 0.506 (0.2350) & 0.449 (0.2378) \\
DeepTTA & 0.829 (0.0310) & \underline{0.820 (0.0458)} & 0.770 (0.0446) & 0.528 (0.1083) & 0.533 (0.2077) & 0.274 (0.1273) &  \underline{0.580 (0.0891)} & 0.562 (0.2681) & 0.409 (0.2527) \\
\midrule
DRPreter & 0.821 (0.0346) & 0.805 (0.0428) & 0.744 (0.0335) & 0.481 (0.0413) & 0.496 (0.2678) & 0.321 (0.1363) & 0.500 (0.1066) & 0.529 (0.2520) & 0.420 (0.2082) \\
Precily & 0.811 (0.0243) & 0.782 (0.0450) & 0.733 (0.0209) & 0.532 (0.0727) & 0.526 (0.2250) & 0.359 (0.2138) & 0.533 (0.0513) & 0.520 (0.2453) & \textbf{0.649 (0.1805)} \\
\midrule
GeneFormer+ECFP & 0.830 (0.0470) & 0.808 (0.0485) & 0.765 (0.0711) & 0.521 (0.1190) & 0.519 (0.2700) & 0.387 (0.1021) & 0.538 (0.1014) & \underline{0.577 (0.2624)} & 0.483 (0.2079)\\
CSG$^2$A & \underline{0.832 (0.0441)} & 0.806 (0.0724) & 0.762 (0.0414) & 0.522 (0.1006) & 0.529 (0.2355) & \textbf{0.603 (0.2299)} & 0.561 (0.1491) & 0.542 (0.2774) & 0.607 (0.1889) \\
 \midrule
PREDIKTOR & \textbf{0.837 (0.0338)} & \textbf{0.829 (0.0491)} & \underline{0.773 (0.0308)} & \textbf{0.594 (0.0900)} & \textbf{0.565 (0.2517)} & 0.553 (0.1985) & \textbf{0.603 (0.0848)} & \textbf{0.595 (0.2837)} & \underline{0.615 (0.1571)} \\
\bottomrule
\end{tabular}
}
\label{tab:main_performance}
\end{table*}

Using the TCGA dataset, we evaluated PREDIKTOR, our proposed multi-view alignment framework, under three distinct cross-validation settings, namely patient-split, drug-split, and tissue-split. Each split setting tests the model's generalization capability differently: patient-split assesses generalization to new patients, drug-split evaluates performance on unseen drugs, and tissue-split examines generalization across distinct cancer types, respectively.

Table \ref{tab:main_performance} summarizes the predictive performance across the three evaluation scenarios. In the patient-split setting, our model achieved an average AUROC of 0.837 (±0.0338), an AUPRC of 0.829 (±0.0491), and  an accuracy of 0.773 (±0.0308). Although PREDIKTOR outperformed all competing methods, other models also achieved strong performance in this setting, with AUROC values exceeding 0.8, suggesting that generalization to unseen patients is relatively less challenging. In the tissue-split setting, our model's performance declined to an AUROC of 0.603 (±0.0848) and an AUPRC of 0.595 (±0.2837), reflecting the increased difficulty of generalizing across cancer types. By contrast, most competing models experienced a more pronounced degradation, with AUROC values falling below 0.6, whereas PREDIKTOR maintained a clear performance margin. The most challenging scenario was the drug-split setting, where PREDIKTOR showed a substantial performance drop (AUROC 0.594 ± 0.09; AUPRC 0.565 ± 0.2517). Under this difficult scenario, knowledge-guided and transformer-based models performed the worst, while simpler traditional machine learning and conventional deep learning models demonstrated comparatively stable behavior. Despite this, PREDIKTOR consistently achieved the best performance across most metrics.

Overall, our model achieved the strongest performance across most metrics in all three split settings, demonstrating particularly strong robustness in the challenging drug-split scenario. This advantage arises from the multi-view design of PREDIKTOR, which effectively integrates complementary strengths of the comparison models in terms of both predictive accuracy and robustness.

\subsubsection{3.1.2 Ablation study of model components}
\begin{figure}[t]
    \centering
    \includegraphics[width=\linewidth]{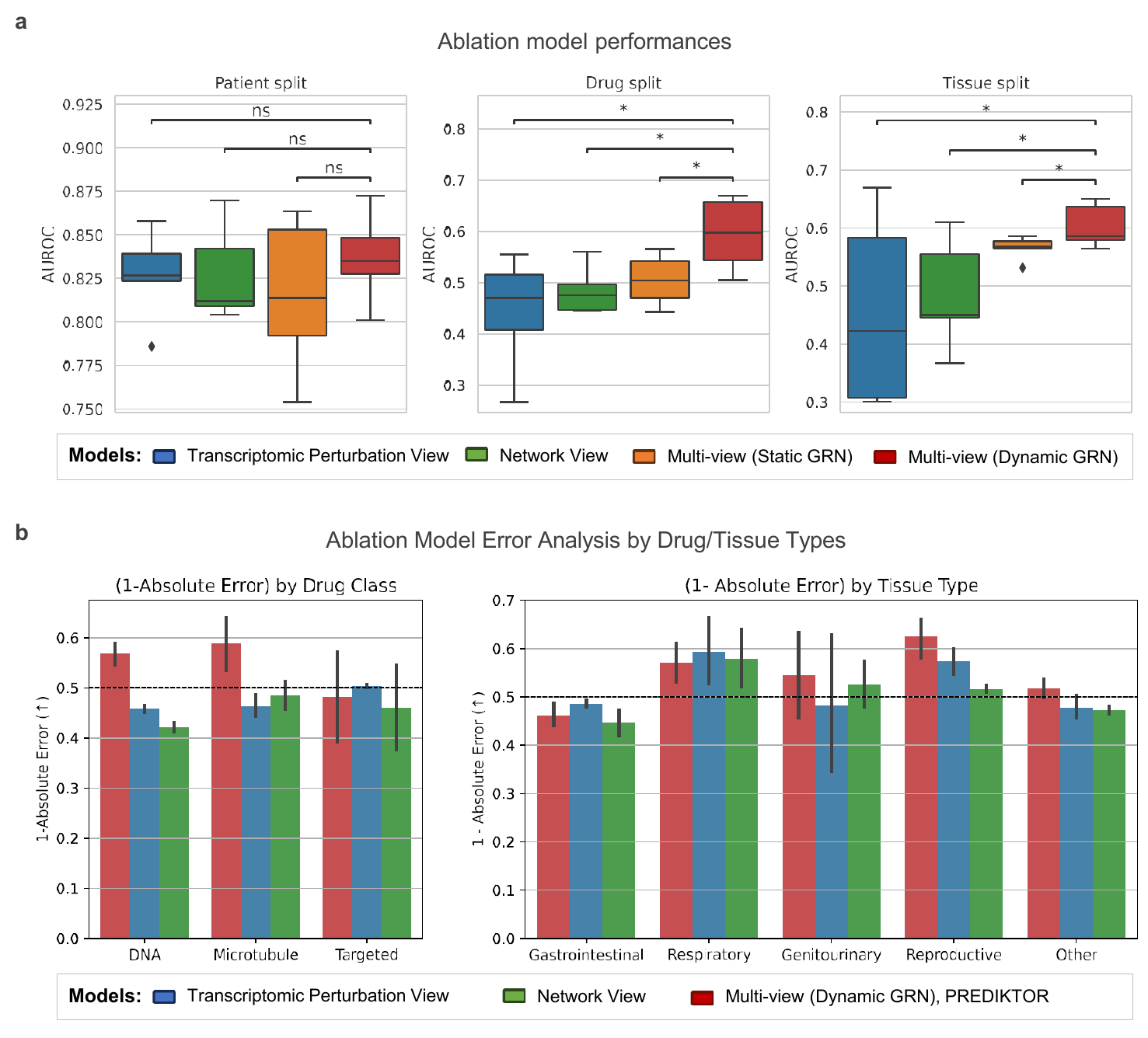}
    \caption{\normalfont \textbf{Performances of ablation models of PREDIKTOR.} \textbf{a} Boxplot of AUROCS for 5-fold CV are provided.  (ns: paired t-test $p$-value $\geq$ 0.05, *: paired t-test $p$-value $<$ 0.05). \textbf{b} Error analysis of each variant models on drug/tissues types. $1-$Absolute Error values are provided, larger the better. The bar plots indicate the mean value and whiskers indicate 95\% Confidence Intervals.}
    \label{fig:ablation_auroc}
\end{figure}

We further conducted an ablation study (Figure \ref{fig:ablation_auroc}a) to dissect the contributions of individual modules in PREDIKTOR. 

Using only the perturbation module led the largest decline in AUROC, with decrease of 0.151 and 0.146 under the drug-split and tissue-split settings, respectively. With the model with only knowledge graph module exhibited smaller AUROC reductions of 0.109 in the drug-split setting and 0.118 in the tissue-split setting. These results indicate that the knowledge graph context contributes more to model generalizability than the perturbation module.
Replacing patient-specific GRN construction with a static GRN topology shared across all patients also resulted in a significant reduction in AUROC under both the drug-split and tissue-split settings, underscoring the importance of modeling individual patient biology. Although the performance decline was smaller than that observed when removing an entire module (0.089 in drug-split and 0.038 in tissue-split setting;s), the decline was still statistically significant (paired t-test, p $<$ 0.05).

We further analyzed the model predictions by drug class and tissue type to characterize error patterns. 
Specifically, drugs in the TCGA dataset were categorized into three classes: DNA-targeting agents, microtubule-targeting agents, and targeted therapies. Tissue types were grouped into five categories: gastrointestinal, respiratory, genitourinary, reproductive, and others. The prediction error was quantified as $1-\text{Absolute Error}$, where the Absolute Error was defined as $|$Predicted probability - Target label$|$. As shown in Figure \ref{fig:ablation_auroc}b, PREDIKTOR, indicated by the red-colored bar, achieves the highest scores in the majority of categories, ranking first in 2 out of 3 drug classes and 3 out of 5 tissue types.

Furthermore, an investigation of PREDIKTOR’s algorithmic components revealed the impact of hard negatives and contrastive learning on prediction performance (Table \ref{tab:ablation_technical}). The PREDIKTOR model, which incorporates hard negatives within the contrastive learning framework, consistently achieved the highest patient drug response prediction performance across all three data-splitting settings. In contrast, removing hard negatives led to a pronounced performance drop on drug-split, yielding results comparable to those obtained by removing contrastive learning, which highlights the essential role of hard negatives in the contrastive learning process. 

Collectively, these results highlight the synergistic benefits of PREDIKTOR’s design, which aligns a patient-specific network view with a transcriptomic perturbation view to enable robust and accurate patient drug response prediction.
\begin{table*}[t]
\centering
\caption{\normalfont \textbf{Evaluation of ablation on technical components.} Patient drug response prediction performances of the proposed PREDIKTOR model along with its variats trained without hard negatives or without contrastive learning, evaluated under three different settings of the TCGA dataset. Mean and standard deviation of 5-fold cross validation are provided.}
\resizebox{0.8\textwidth}{!}{%
\begin{tabular}{l|rr|rr|rr}
\toprule
 & \multicolumn{2}{c|}{\textbf{Patient split}} & \multicolumn{2}{c|}{\textbf{Drug split}} & \multicolumn{2}{c}{\textbf{Tissue split}} \\
 & \multicolumn{1}{c}{\textbf{AUROC}} & \multicolumn{1}{c|}{\textbf{AUPRC}} & \multicolumn{1}{c}{\textbf{AUROC}} & \multicolumn{1}{c|}{\textbf{AUPRC}} & \multicolumn{1}{c}{\textbf{AUROC}} & \multicolumn{1}{c}{\textbf{AUPRC}} \\
 \midrule
PREDIKTOR & \textbf{0.837 (0.0338)} & \textbf{0.829 (0.0491)} & \textbf{0.594 (0.0900)} & \textbf{0.565 (0.2517)} & \textbf{0.603 (0.0848)} & \textbf{0.595 (0.2837)} \\
w/o hard negative& 0.834 (0.0425) & 0.816 (0.0643) & 0.574 (0.0394) & 0.560 (0.2510) & 0.592 (0.0796) & 0.588 (0.2782) \\
w/o contrastive& 0.832 (0.0368) & 0.808 (0.0531) & 0.559 (0.0310) & 0.541 (0.2518) & 0.540 (0.0214) & 0.553 (0.2652)\\
\bottomrule
\end{tabular}
}
\label{tab:ablation_technical}
\end{table*}

\subsection{3.2 Generalization to an external clinical dataset}

\begin{table*}[t]
\centering
\caption{\normalfont \textbf{Evaluation of generalizability of comparison models.} Patient drug response prediction performances on the external clinical dataset (I-SPY2 Trial). Models trained on TCGA-Tissue split dataset are utilized in a zero-shot setting (without further fine-tuning). Mean and standard deviation of 5-fold cross validation are provided. Best performances are marked in bold and second-best are underlined.}
\resizebox{0.6\textwidth}{!}{%
\begin{tabular}{lcc|rrr}
\toprule
 & \multirow{2}{*}{\begin{tabular}[c]{@{}c@{}} Knowledge Graph\\Information\end{tabular}} & \multirow{2}{*}{\begin{tabular}[c]{@{}c@{}}Gene-level \\Transfer\end{tabular}}& \multicolumn{3}{c}{\textbf{External Dataset (I-SPY2 Trial)}} \\
 &  &  & \multicolumn{1}{c}{AUROC} & \multicolumn{1}{c}{AUPRC} & \multicolumn{1}{c}{F1 score} \\
 \midrule
XGBoost & \xmark & \xmark & 0.503 (0.0049) & 0.169 (0.0024) & 0.115 (0.1406) \\
Random Forest & \xmark & \xmark& 0.492 (0.0131) & 0.167 (0.0024) & 0.167 (0.0024) \\
SVM & \xmark & \xmark &  0.516 (0.0737) & 0.183 (0.0256) & 0.000 (0.0000) \\
DeepCDR & \xmark & \xmark & 0.524 (0.0606) & 0.207 (0.0421) & 0.230 (0.1148) \\
DeepTTA & \xmark & \xmark & 0.559 (0.0756) & 0.219 (0.0771) & 0.172 (0.1406) \\
\midrule
DRPreter & \cmark & \xmark & 0.457 (0.0602) & 0.174 (0.0388) & 0.199 (0.0990)\\
Precily & \cmark & \xmark & 0.480 (0.0468) & 0.178 (0.0207)& 0.167 (0.0678)  \\
\midrule
GeneFormer+ECFP & \xmark & \cmark & \underline{0.626 (0.0811)} & \underline{0.271 (0.0640)}  & 0.225 (0.1131) \\
CSG$^2$A & \xmark & \cmark & 0.568 (0.0802) & 0.232 (0.0598) & \underline{0.287 (0.0000)} \\
\midrule
PREDIKTOR & \cmark & \cmark &  \textbf{0.661 (0.0429)} & \textbf{0.296 (0.0499)} & \textbf{0.306 (0.0262)} \\
\bottomrule
\end{tabular}%
}
\label{tab:ispy_performance}
\end{table*}

To rigorously assess translational generalization, we applied the trained models in a zero-shot inference setting to an external, clinically derived dataset (I-SPY2\cite{barker2009ispy}), consisting of breast cancer patients treated exclusively with paclitaxel. In this setting, PREDIKTOR achieved the best performance among all evaluated methods, with an AUROC of 0.661, an AUPRC of 0.296, and an F1 score of 0.306. As summarized in Table \ref{tab:ispy_performance}, it consistently outperformed traditional machine-learning and deep-learning baselines, as well as knowledge-guided and transfer learning based attention models, which showed limited generalization under zero-shot conditions. Although transfer learning based attention models achieved the second-best average performance, their AUROC standard deviation was substantially higher (approximately 0.08) than that of other baselines and knowledge-guided models, and nearly twice that of PREDIKTOR (0.0429), indicating less stable performance.

Collectively, these results demonstrate both the strong predictive accuracy and robustness of PREDIKTOR for unseen clinical drug response prediction.

\subsection{3.3 Biological interpretability and mechanistic relevance of PREDIKTOR in gene-level explanations}

\begin{figure}[!t]
    \centering
    \includegraphics[width=\linewidth]{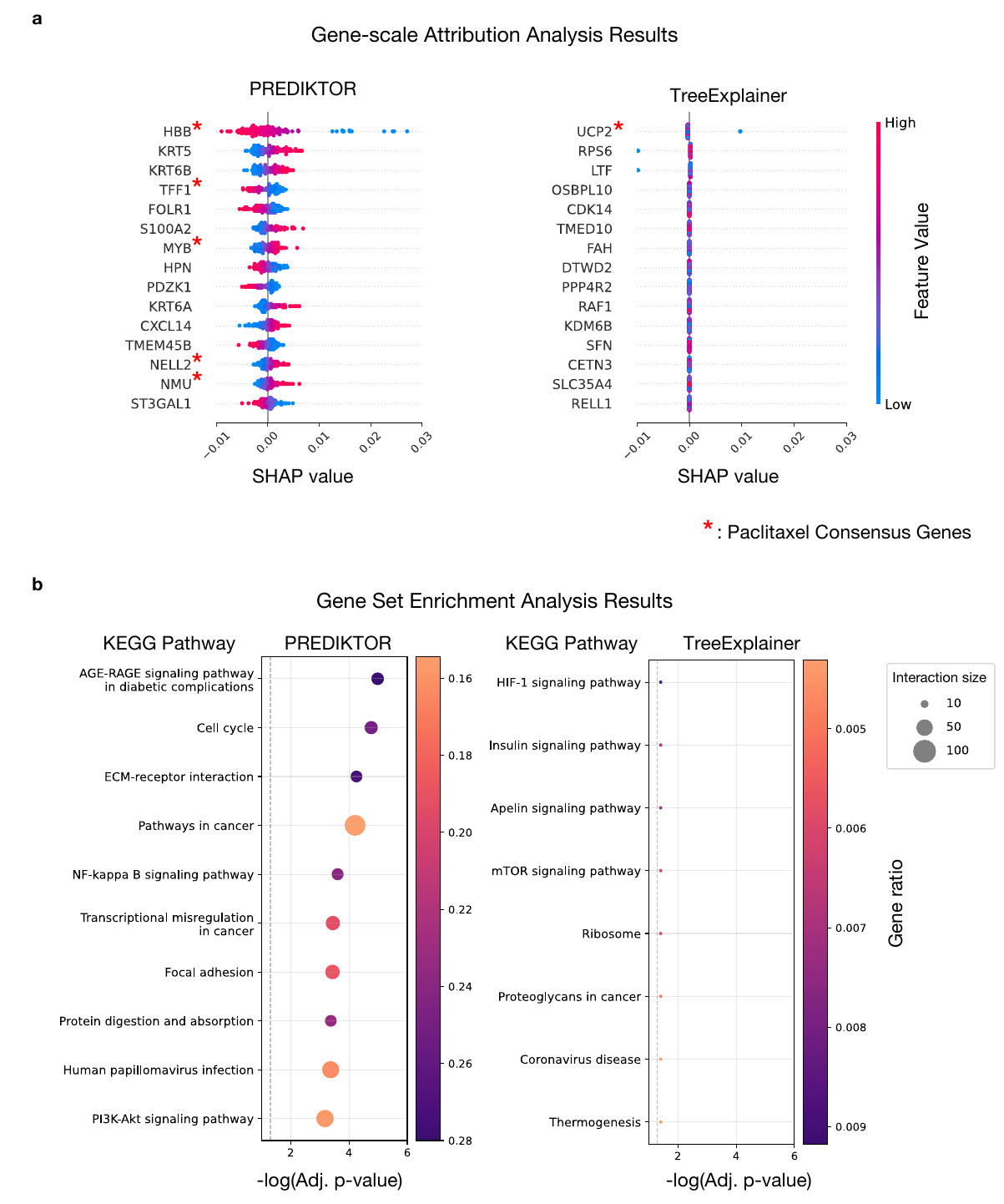}
    \caption{\normalfont \textbf{Comparative gene attribution and pathway enrichment analysis.}
    \textbf{a} Gene-level attribution profiles are obtained from PREDIKTOR using expected gradients and from Random Forest baseline using SHAP values computed with TreeExplainer. \textbf{b} Gene set enrichment analysis (GSEA) performed on the top-ranked genes identified by PREDIKTOR and TreeExplainer.
    }
    \label{fig:prediktor_interpretation}
\end{figure}

We assessed interpretability by comparing our multi-view model with TreeExplainer \cite{lundberg2020local}, a SHAP-based method \cite{lundberg2017unified}. A Random Forest trained on TCGA paclitaxel-treated breast cancer transcriptomes served as the baseline. TreeExplainer was used to compute SHAP values for gene attribution in the baseline model, whereas we applied an explainable gradients approach, analogous to SHAP but optimized for deep neural networks, to interpret predictions from our trained MLP classifier. External validation was conducted on the I-SPY2 cohort, selecting the top 10\% of genes with the highest absolute attribution scores.

Figure \ref{fig:prediktor_interpretation}a illustrates the gene-level attribution profiles obtained from PREDIKTOR using expected gradients and from the Random Forest baseline using TreeExplainer. PREDIKTOR distributes attribution more broadly across genes, reflecting contributions from a diverse set of molecular signals. In contrast, TreeExplainer assigns attribution to a more limited subset of genes, which may underrepresent relevant biological pathways. This difference suggests that PREDIKTOR provides a more comprehensive and interpretable characterization of gene contributions, consistent with the multifactorial nature of biological processes.

To further assess the biological relevance of the model-derived gene attributions, we conducted gene set enrichment analysis (GSEA) on the top-ranked genes. As shown in Figure \ref{fig:prediktor_interpretation}b, TreeExplainer yielded eight KEGG pathways with significant enrichment (adjusted $ P <$0.05); however, none of these pathways demonstrated known mechanistic connections to paclitaxel response or breast cancer biology. In contrast, our fused multi-view framework identified a suite of pathways that are not only statistically significant but also mechanistically aligned with established pharmacological and oncological knowledge. Among the top 10 enriched pathways were those directly implicated in paclitaxel’s mechanism of action, including \textit{Proteoglycans in cancer}, \textit{Cell cycle}, and \textit{Chemokine signaling pathway} \cite{zhao2022mechanisms}, as well as hallmark pathways in breast cancer biology, such as \textit{p53 signaling} and \textit{PI3K–Akt signaling} \cite{dai2016cancer}. Togehter, these results underscore the model’s capacity to yield biologically interpretable and therapeutically relevant insights.

\subsubsection{3.3.1 Quantitative validation through cross-referencing with paclitaxel consensus signatures}
To assess the biological validity of PREDIKTOR’s gene-level explanations, we conducted a targeted analysis of SHAP-derived feature attributions for patients treated with paclitaxel. Specifically, we evaluated whether genes with high SHAP values aligned with known mechanistic signatures of paclitaxel sensitivity, as documented in the LINCS L1000 consensus dataset, retrieved through Enrichr \cite{xie2021gene}.
Among the 241 LINCS consensus genes upregulated by paclitaxel treatment, PREDIKTOR identified 64 genes, reflecting a statistically significant enrichment (p-value = 1.4E-7). Additionally, for the 249 genes downregulated by paclitaxel, PREDIKTOR identified 62 genes, further highlighting statistically significant enrichment (p-value = 3.8E-6). Notably, within the top 100 genes ranked by SHAP values, 16 were validated paclitaxel-associated signature genes, clearly demonstrating the mechanistic and functional interpretability of our multimodal approach. A comparable number of validated paclitaxel-associated signature genes was consistently observed across experiments with different random seeds (Table~\ref{tab:supp_seed_consensus_top100}).
In contrast, the TreeExplainer-based baseline identified only one of these validated marker genes (UCP2) in the external cohort, suggesting improved biological relevance achieved through the integration of multi-view, biologically-informed features.

\begin{table*}[]
\centering
\caption{\normalfont \textbf{Comparative evaluation of gene-level semantic information captured by transcriptomic perturbation view and knowledge graph–based network view.}
Prediction performance of Gene Ontology (GO) term classification from gene embeddings derived from the transcriptomic perturbation module (perturbation module) and knowledge graph-based network module (network module) of PREDIKTOR. Across all three GO categories, the network module achieves substantially higher $F_{max}$ and AUPRC scores, indicating that the network module encodes richer functional and contextual information than the purely data-driven perturbation module. Mean and standard deviation of 5-fold cross validation are provided.}
\resizebox{0.85\textwidth}{!}{%
\begin{tabular}{l|rr|rr|rr}
\toprule
 & \multicolumn{2}{c|}{\textbf{GO Molecular   Function}} & \multicolumn{2}{c|}{\textbf{GO Biological   Process}} & \multicolumn{2}{c}{\textbf{GO Cellular   Component}} \\
\textbf{Gene Representation} & \multicolumn{1}{c}{Fmax} & \multicolumn{1}{c|}{AUPRC} & \multicolumn{1}{c}{Fmax} & \multicolumn{1}{c|}{AUPRC} & \multicolumn{1}{c}{Fmax} & \multicolumn{1}{c}{AUPRC} \\
\midrule
Perturbation module& 0.097 (0.011) & 0.028 (0.004) & 0.043 (0.004) & 0.009 (0.001) & 0.276 (0.011) & 0.128 (0.007) \\
Network module& 0.167 (0.019) & 0.073 (0.009) & 0.086 (0.007) & 0.028 (0.007) & 0.380 (0.013) & 0.274 (0.019) \\
\bottomrule
\end{tabular}%
}\label{tab:GOpred}
\end{table*}

\subsubsection{3.3.2 Gene-level functional information encoded by model modules}
To evaluate whether the knowledge graph-based network view (network module) and transcriptomic perturbation view (perturbation module) capture distinct aspects of biological information at the gene representation level, we conducted a functional annotation prediction task using Gene Ontology (GO) terms. Specifically, we assessed how well embeddings from each module encode gene-level semantic information across the three primary GO domains: Molecular Function (MF), Biological Process (BP), and Cellular Component (CC). For each module, the learned gene embeddings were extracted and used as input features to train a multi-label logistic regression classifier for GO term prediction. Predictive performance was quantified using the maximum F-measure ($F_{max}$) and the AUPRC, averaged across cross-validation folds.

As shown in Table \ref{tab:GOpred}, embeddings from the network module consistently outperform those from the perturbation module across all GO categories. The KG-based network module achieve markedly higher $F_{max}$ and AUPRC scores, improving by roughly 1-2 folds, depending on the ontology domain, indicating that they encode richer and more functionally relevant biological information. These results support the hypothesis that mechanistic priors enhance biological coherence in learned representations. While the perturbation module captures statistical dependencies derived from large-scale perturbation data, the network module integrates topological and relational context inherent to molecular interaction networks, yielding more interpretable and biologically informed embeddings.

Overall, our results demonstrate that integrating knowledge graph and perturbation-based representations substantially improves the biological interpretability of AI-driven drug response prediction, facilitating the identification of clinically actionable biomarkers relevant to paclitaxel sensitivity and breast cancer biology. 

\section{4 Discussion}

PREDIKTOR achieved overall improvements over existing baselines across patient-, drug-, and tissue-split settings, as well as in the external I-SPY2 validation cohort. These results suggest that integrating patient-specific regulatory context with transferable perturbation information can improve personalized drug response prediction from pre-treatment transcriptomes. In particular, the ablation studies support the complementary roles of the two views: the network view captures individualized disease-specific regulatory structure, whereas the perturbation view provides transferable perturbation priors learned from large-scale perturbation data. In addition, we kept the pretrained CSG$^2$A and MAT modules frozen to preserve transferable perturbation priors and reduce overfitting to the limited TCGA cohort, consistent with prior observations that fine-tuning CSG$^2$A can degrade downstream performance \cite{bang2024transfer}.

Despite these promising results, several limitations remain. First, although PREDIKTOR outperformed baselines in the drug-split setting, its absolute performance was lower than in the patient-split setting, indicating that generalization to entirely unseen compounds remains challenging. This limitation may arise because the current perturbation view primarily relies on SMILES-based molecular structure representations, while the network view depends on known drug-target interactions from DrugBank. For cold-start or investigational drugs without documented target genes, the network view is not directly applicable as intended because the drug node cannot be connected to the patient-specific regulatory network through drug-target edges. In such cases, the loss of network-view information is expected to reduce both predictive performance and mechanistic interpretability, particularly for drugs whose biological effects cannot be adequately inferred from chemical structure alone.

Future work should therefore focus on improving novel-drug generalization and reducing dependence on incomplete prior knowledge. One promising direction is to infer provisional drug-target edges using chemical similarity, drug-target interaction prediction, or chemoproteomic evidence, and incorporate these inferred links into the patient-specific network view. In addition, the perturbation view could be further enhanced with richer drug representations beyond SMILES, such as 3D molecular features, LLM-derived drug descriptions, predicted drug-target interaction profiles, and pathway-level activity profiles. Future work could also explore more adaptive fusion strategies, such as attention-based fusion, gated fusion, or cross-view interaction modules, to better capture context-dependent relationships between patient-specific regulatory structure and drug-induced perturbation patterns. These extensions may improve the model’s ability to capture downstream drug effects and generalize to novel compounds.

\section{5 Conclusion}
In this study, we present a multi-view framework designed to improve personalized drug response prediction by integrating patient-specific network view with transferable transcriptomic perturbation view. Motivated by the challenges of predicting clinical drug responses directly from patient molecular profiles,  we propose a framework that enhances both predictive performance and interpretability in precision oncology.

Our method integrates two complementary views of a patient. First, we construct personalized GRNs for each patient using DysRegNet and augment it with drug–target links from DrugBanK to form a knowledge graph-based network view. Second, we incorporate a condition-specific gene-gene attention module (CSG$^2$A), pretrained on large-scale perturbation response data, to derive a transcriptomic perturbation view. The representations from these two views are then fused within a contrastive learning framework, where hard negative samples are defined as representations of the same patient under different drug treatments. 

We evaluate our framework on the TCGA dataset under patient-, drug-, and tissue-split scenarios, and further test its generalization to the I-SPY2 clinical trial dataset in a zero-shot inference setting. Across all settings, our model consistently outperforms baselines, achieving state-of-the-art performance. Ablation studies confirm the contributions of personalized knowledge graph-based network view and transciptomic perturbation view, as well as the importance of their alignment. Moreover, qualitative analyses demonstrate that our model’s predictions are consistent with  known biological mechanisms, providing  biologically informed explanations that support downstream experimental validation and clinical interpretation. 

While showing promising outcomes, our work has several limitations. Training and validation datasets, particularly the TCGA, reflect a limited range of cancer types and drug exposures, which may restrict generalizability. In addition, the biological knowledge graphs used remain incomplete and are limited to known interactions. Future extensions could include the integration of additional molecular modalities such as protein-level data, spatial transcriptomics, or single-cell sequencing to refine the resolution of patient models.

In summary, our results demonstrate that the alignment of patient-specific knowledge graph-based network view with the transcriptomic perturbation view enables both accurate and interpretable drug response prediction in oncology. We expect this framework to generalize to other clinical prediction tasks that require personalized prediction and mechanistic insight, contributing to the development of transparent and clinically actionable machine learning approaches. 

\section*{Code availability}

The source code for model training and inference, together with configuration files and trained model parameters, will be made available upon request.

\section{Author contributions statement}
D.B., I.S. and I.Y. conducted the experiments, and D.B., S.A., I.Y., and S.L. wrote the manuscript. All authors conceived the experiments, analyzed the results, and reviewed the manuscript. 

\section{Acknowledgments} 
This work was supported by the INHA UNIVERSITY Research Grant, the National Research Foundation of Korea (NRF) grants funded by the Korean government (MSIT) (RS-2026-25473777, RS-2026-25518860, RS-2023-NR077172, and RS-2024-00431505), the Technology Innovation Program (RS-2025-13642970, Development of an AI-based Platform for Predicting and Evaluating Drug Safety and Efficacy) funded by the Ministry of Trade, Industry and Resources (MOTIR, Korea), and a grant (RS-2026-25506298) from Ministry of Food and Drug Safety in 2026-2030. This study was also funded by AIGENDRUG Co., Ltd.





\bibliographystyle{unsrt}
\bibliography{references}

\clearpage
\onecolumn
\appendix
\section*{Supplementary Material}
\setcounter{figure}{0}
\renewcommand{\thefigure}{S\arabic{figure}}
\setcounter{table}{0}
\renewcommand{\thetable}{S\arabic{table}}

\begin{figure}[!htbp]
    \centering
    \includegraphics[width=\textwidth,height=0.82\textheight,keepaspectratio]{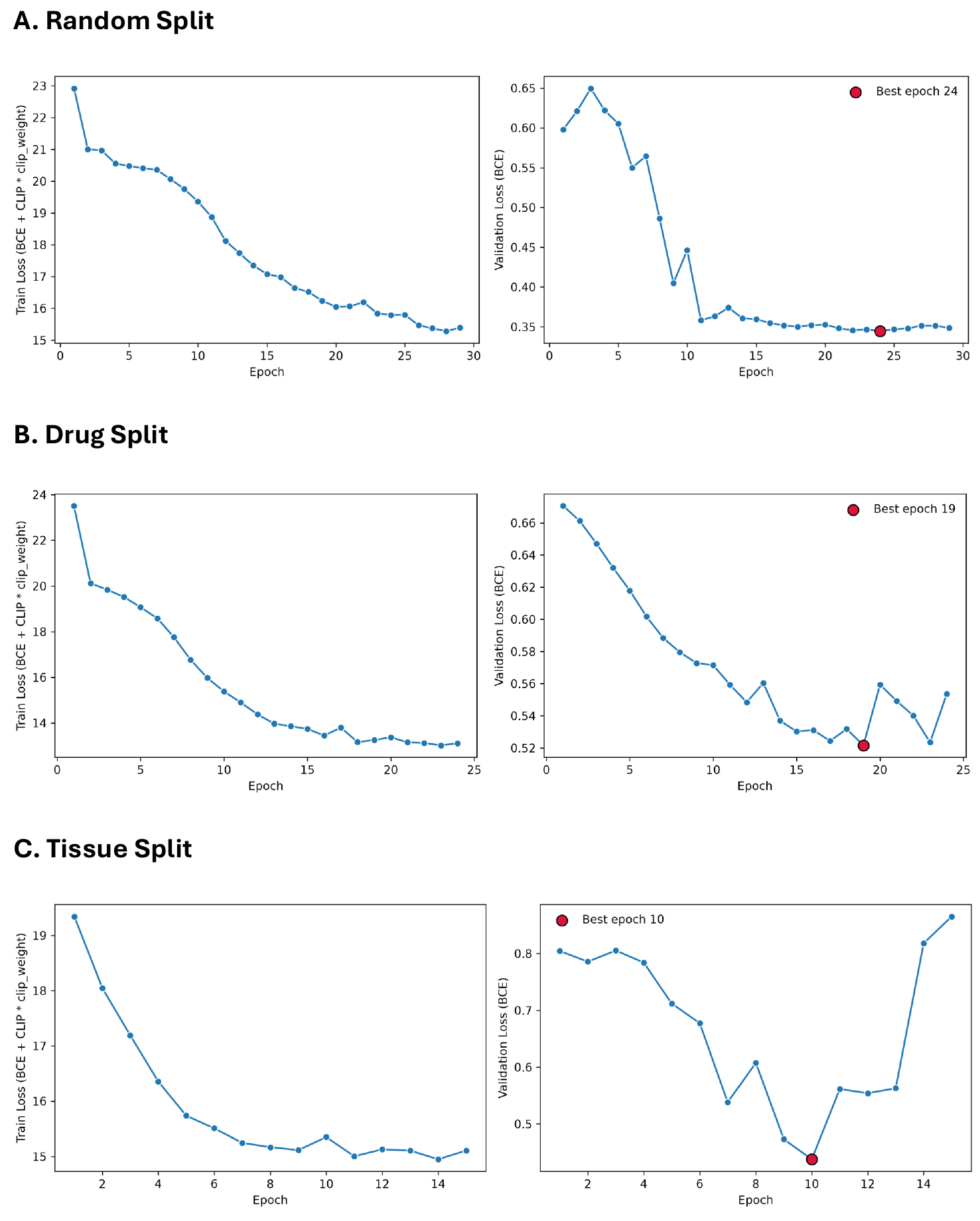}
    \caption{\normalfont \textbf{Representative training and validation loss curves.}
    Training and validation loss trajectories of PREDIKTOR from one representative cross-validation fold under the random-, drug-, and tissue-split settings. The curves illustrate model convergence across epochs under each data-splitting scenario. Red markers indicate the epochs with the lowest validation loss, which were selected as the best checkpoints for the corresponding runs. 
    }
    \label{fig:supp_lossplot}
\end{figure}


\begin{table*}[p]
\centering
\caption{\normalfont \textbf{Stability of top-100 PREDIKTOR gene attributions across random initializations.}
For each random-seed replicate trained using the same tissue-split partition, genes were ranked according to their expected-gradients attribution scores for I-SPY2 paclitaxel samples. The table reports the number of genes among the top 100 that overlapped with the Enrichr LINCS L1000 Paclitaxel Up, Paclitaxel Down, and combined consensus signatures.}
\label{tab:supp_seed_consensus_top100}

\setlength{\tabcolsep}{18pt}

\begin{tabular}{c|ccc}
\toprule
\textbf{Random Seed} & \textbf{Paclitaxel Up} & \textbf{Paclitaxel Down} & \textbf{Union} \\
\midrule
101 & 12 & 5 & 17 \\
102 & 9 & 5 & 14 \\
103 & 10 & 4 & 14 \\
104 & 13 & 4 & 17 \\
105 & 9 & 3 & 12 \\
\bottomrule
\end{tabular}

\end{table*}

\end{document}